\documentclass[10pt,twocolumn,letterpaper]{article}

\usepackage[pagenumbers]{wacv} 

\usepackage{multirow}

\providecommand{\Description}[1]{}

\definecolor{wacvblue}{rgb}{0.21,0.49,0.74}
\usepackage[pagebackref,breaklinks,colorlinks,allcolors=wacvblue]{hyperref}

\title{FairLRF: Achieving Fairness through Sparse Low Rank Factorization}

\author{Yuanbo Guo\\
University of Notre Dame\\
Notre Dame, Indiana, USA\\
\and
Jun Xia\\
University of Notre Dame\\
Notre Dame, Indiana, USA\\
\and
Yiyu Shi\\
University of Notre Dame\\
Notre Dame, Indiana, USA\\
}

\begin{document}
\maketitle
\begin{abstract}
As deep learning (DL) techniques become integral to various applications, ensuring model fairness while maintaining high performance has become increasingly critical, particularly in sensitive fields such as medical diagnosis.
Although a variety of bias-mitigation methods have been proposed, many rely on computationally expensive debiasing strategies or suffer substantial drops in model accuracy, which limits their practicality in real-world, resource-constrained settings.
To address this issue, we propose a fairness-oriented low rank factorization (LRF) framework that leverages singular value decomposition (SVD) to improve DL model fairness.
Unlike traditional SVD, which is mainly used for model compression by decomposing and reducing weight matrices, our work shows that SVD can also serve as an effective tool for fairness enhancement.
Specifically, we observed that elements in the unitary matrices obtained from SVD contribute unequally to model bias across groups defined by sensitive attributes.
Motivated by this observation, we propose a method, named FairLRF, that selectively removes bias-inducing elements from unitary matrices to reduce group disparities, thus enhancing model fairness.
Extensive experiments show that our method outperforms conventional LRF methods as well as state-of-the-art fairness-enhancing techniques in terms of improving fairness while maintaining accuracy.
Additionally, an ablation study examines how major hyper-parameters may influence the performance of processed models.
To the best of our knowledge, this is the first work utilizing SVD not primarily for compression but for fairness enhancement of DL classifications.
\end{abstract}
    
\section{Introduction}

Deep learning (DL) has been widely adopted into our daily life with various applications, while fairness problems of DL models have been obtaining growing attention as well.
Biased usage of DL tools has been noted to contribute to systematic discrimination and injustice, and to go against the interests of unprivileged groups of people~\cite{ferrara2024fairness}.
For example, recruiting tools can show different preferences on genders~\cite{dastin2022amazon}, and healthcare commercial algorithm works with biases against certain racial groups~\cite{obermeyer2019dissecting}.
As DL models get more and more involved in making all kinds of decisions, it is of vital importance to promote DL fairness in research as well as applications.

It is noted that there have been a variety of researches regarding fairness, which come up with concepts of fairness based on different views.
For instance, \cite{dunkelau2019fairness} requires a fair classifier to guarantee the  predictions assigning ``favourable outcome'' to individuals of the privileged and unprivileged groups do not discriminate over the protected attribute.
Another work~\cite{du2020fairness} categorizes unfairness into ``prediction outcome discrimination'' and ``prediction quality disparity''.
The former refers to cases when a model produces unfavorable treatment of people due to certain demographic groups, while the latter refers to situations when a model shows lower inference quality for some groups of people as opposed to other groups.
To clarify the research objectives, in this paper, we focus on popular fairness metrics that help us objectively measure the fairness of predictions of a DL model, which will be further discussed later.

Meanwhile, model compression is one of the crucial steps before the deployment and inference of DL models, especially for DL applications on devices with limited resource capability.
Typical model compression methods include pruning, quantization, low rank factorization (LRF), and so on.
These tools focus on the trade-off between accuracy and efficiency (e.g., inference speed), yet there has been a few researches that surprisingly utilize compression methods towards the fairness improvement objective~\cite{guo2024hardware}.
Multiple works~\cite{wu2022fairprune,chiu2023toward,kong2024achieving} involve pruning, a technique that shrinks model size and increases inference speed by removing partial weights from a DL model, and~\cite{guo2024fairquantize} uses quantization which reduces space to store weights to benefit efficiency.
Both routes improve fairness by tuning the compression progress towards fairness improvement instead of merely maintaining accuracy performance like conventional solutions.
It is noticed, however, that no such research focuses on fairness enhancement through LRF so far.
Singular value decomposition (SVD) is one of popular ways to LRF, which significantly reduces the number of weights without sacrificing accuracy by dropping less significant components after decomposing weight matrices.
\cite{swaminathan2020sparse} proposed a method to further increase sparsity after applying SVD, but it only paid attention to improving compression rate while minimizing accuracy loss.

Therefore, in this paper, a novel framework, named FairLRF, is proposed to address this gap.
It mitigates biases by tuning the original LRF, a method previously used for model compression.
To achieve this, we measure the difference of weights in target matrices in terms of how much they contribute to prediction biases on different sensitive attributes.
Then based on such information, weights that contribute more to unfairness can get dropped while others with less influence get preserved.
Experiments are conducted to compare our work with conventional LRF methods as well as other fairness-oriented compression methods with various setups and datasets to cover different situations of DL applications.
The results show that FairLRF is capable of achieving a better balance between accuracy and fairness while requiring less effort.
The source code for FairLRF is publicly available upon acceptance.

\section{Method}

\begin{figure*}[ht]
  \centering
  \includegraphics[width=0.7\textwidth]{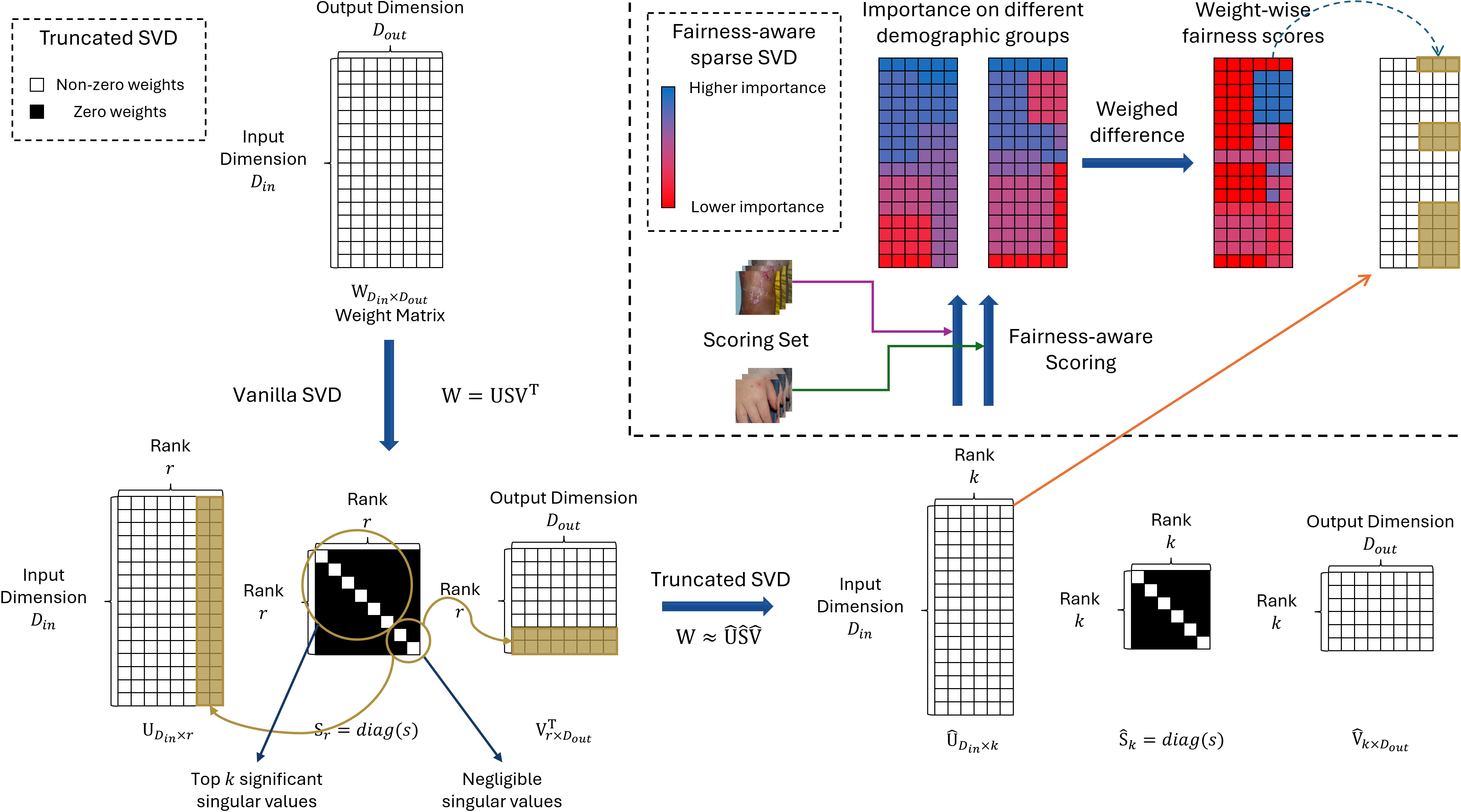}
  \caption{
  Workflow of FairLRF.
  Truncated SVD demonstrations are originated from~\cite{swaminathan2020sparse}; scoring set image samples come from Fitzpatrick-17k.
  The importance values during sparse SVD are randomly filled, and numbers like weight matrix sizes are for demonstrative purposes only.
  The sparse SVD process for the other unitary matrix is omitted for simplicity.
  }
  \Description{
  Workflow of FairLRF.
  Truncated SVD demonstrations are originated from~\cite{swaminathan2020sparse}; scoring set image samples come from Fitzpatrick-17k.
  The importance values during sparse SVD are randomly filled, and numbers like weight matrix sizes are for demonstrative purpose only.
  The sparse SVD process for the other unitary matrix is omitted for simplicity.
  The most important step is to measure weight-wise importance of matrices on two subsets of sample data grouped by the sensitive attribute.
  }
  \label{workflow}
\end{figure*}

\subsection{Problem Definition and Fairness Metrics}

As introduced above, it is necessary to first clearly determine what the fairness objective is in this work.
For a classification dataset $D = \{(x_i, y_{0i}, c_i)\}, i \in \{1, 2, ..., N\}$, each data point consists of the input $x_i$, ground truth for the target attribute $y_{0i}$ (or class label), and the sensitive attribute $c_i$.
The sensitive attribute depends on specific cases, and it can be gender, skin tone, age, etc.
Suppose we already have a classification model pre-trained on the training set of the dataset, denoted as $y = F(\Theta, x)$, where $\Theta$ is the weights of the model $F$, then we can naturally consider to improve fairness by reducing the bias of the model prediction performance on different groups divided by the sensitive attribute $c$.
In this paper, we focus on binary sensitive attributes ($c_i \in \{0, 1\}$), thus each data point can be divided into the privileged group (with better performance of the pre-trained model) and the unprivileged group (with worse performance, i.e., the pre-trained model shows discrimination against this group).

To quantitatively measure how fair (or biased) the model is, we use equalized opportunity and equalized odds~\cite{hardt2016equality} in this work, which have been widely used in previous research regarding fairness.
For equalized opportunity, it is the difference of true negative rates (denoted as EOpp0) or true positive rates (denoted as EOpp1) of different demographic groups within each class of target attributes, followed by sum of them.
For equalized odds (denoted as EOdd), it collects the difference of true positive rates plus difference of false positive rates.
They can be written into the following equations:
\begin{equation}
    EOpp0 = \sum_{k=1}^{K}{ \lvert TNR_k^1 - TNR_k^0 \rvert },
\end{equation}
\begin{equation}
    EOpp1 = \sum_{k=1}^{K}{ \lvert TPR_k^1 - TPR_k^0 \rvert },
\end{equation}
\begin{equation}
    EOdd = \sum_{k=1}^{K}{ \lvert TPR_k^1 - TPR_k^0 + FPR_k^1 - FPR_k^0 \rvert }.
\end{equation}
For all of EOpp0, EOpp1, and EOdd, the smaller the better.

These metrics reflect the statistical differences of model performance on different demographic groups.
Given that fairness for positive cases is usually more important in real world scenarios like disease diagnosis, which will be involved in experiments later, we use EOpp1 as the representation of equalized opportunity in this paper, denoted as EOpp below.

\subsection{Sparse SVD: Contributions of Weights Differ}

To begin with, for a weight matrix $W_{D_{in} \times D_{Out}}$ of a fully connected layer, we can use SVD to convert it into three matrices, $W = U_{D_{in}} S_{D_{in} \times D_{out}} V_{D_{out}}^\intercal$, where $U$ and $V$ are unitary matrices and $S$ is a diagonal matrix.
The diagonal entries of $S$ is also known as singular values, which are non-negative real numbers and are often ordered in the descending order.
In this way, only the first few singular values show significant contribution to the original matrix, leading to a popular approximation of SVD, truncated SVD.
By preserving only the first $k$ singular values, we truncate the remaining parts of $S$ and sequentially truncate the unitary matrices, getting an approximation $W \approx \hat{U}_{D_{in} \times k} \hat{S}_{k} \hat{V}_{k \times D_{out}}$ (dropping transpose for simplicity).

Previous researches~\cite{lebedev2018speeding,cheng2017survey} revealed that vanilla DL models often consist of redundant neurons with different contributions towards the outputs.
This inspired not only weight-removal techniques like pruning to remove select parts of models, but also LRF methods to take a step further.
As introduced in~\cite{swaminathan2020sparse}, it is also possible to further drop some weights from truncated unitary matrices to achieve even higher compression rate after truncated SVD while minimizing the influence on performance.
Consider the truncated SVD $W \approx \hat{U}_{D_{in} \times k} \hat{S}_{k} \hat{V}_{k \times D_{out}}$ introduced above.
For one truncated unitary matrix $\hat{U}_{D_{in}}$, we can apply sparsity to some of its rows, and the ratio of rows to get sparse is \textbf{sparsity rate} $sr \in [0, 1]$, i.e., there are $sr \times D_{in}$ rows getting sparse.
Next, a certain portion of weights at the end of rows (matching to the least significant few singular values in $\hat{S}_{k}$) gets zeroed out, where the ratio of the remaining weight number and original rank is \textbf{reduction rate} $rr \in [0, 1]$.
In other words, there will be $\lfloor rr \times k \rfloor$ weights kept as a reduced representation, and $\lceil (1.0 - rr) \times k \rceil$ weights dropped.
Similarly, each of $sr \times D_{out}$ selected columns in $\hat{V}_{k \times D_{out}}$ drops all but $rr \times k$ weights.
Known as sparse SVD, this method extends truncated SVD to reduce more weights, and naturally it is of vital importance to determine which strategy is followed during such processes.
However, there were only strategies borrowed from conventional pruning (we will introduce later) in order to determine where the sparsification applies to.
For instance, the absolute weight strategy is to compare the row-wise or column-wise sum of absolute values of the original weight matrix, and to apply sparsity to those with the least sum, as these weights are considered as the least important ones.
Therefore, we propose a new strategy below to guide sparse SVD, not only towards accuracy preservation, but also for fairness improvement.

It is noted that this strategy is not limited to merely fully connected layers.
Instead, it applies to any layer that is essentially doing matrix multiplications between input matrices and weight matrices.
For instance, im2col is one of major techniques to conduct convolution operations in terms of matrix multiplications, which is of vital importance for acceleration of convolutional layers, and there have been many research works~\cite{7995254,9650846} focusing on im2col.
However, convolutional layers tend to have less weights than fully connected layers, as weights in a convolutional layer only come from the convolutional kernels.
Therefore, for simplicity, this paper only discusses the compression of fully connected layers to maximize compression effects on a single layer.

\subsection{Fairness-oriented Sparse SVD}

As it turns out that different contributions of weights matters, we can consider the different importance of weights in terms of fairness instead of mere accuracy to guide sparse SVD to optimize fairness based on previously mentioned observations.
Such reforming of conventional compression methods has been validated to work in recent researches~\cite{wu2022fairprune,guo2024fairquantize,kong2024achieving} as well.

Mathematically, when we ``remove'' a weight from the weight matrix, we are effectively setting its value to zero.
Given a DL model $F$ and its objective function $E$, the change of the objective function after zeroing out one weight $\theta$ from the post-truncated SVD pre-trained weights $\Theta$ can be approximated by a Taylor series~\cite{lecun1989optimal},
\begin{equation}\label{equ:taylor_exp}
\begin{split}
\Delta E & = E(x | \theta_i = 0) - E(x) \\
& = -\sum_i g_i \theta_i + \frac{1}{2}\sum_{i} h_{ii} \theta_i^2 + \frac{1}{2} \sum_{i\neq j}h_{ij}\theta_i \theta_j + O(\theta_i^3) \\
& \approx \sum_i \frac{1}{2} h_{ii} \theta_i^2,
\end{split}
\end{equation}
where $x$ is a sample input.
In the first term, $g_i = \frac{\partial E}{\partial {\theta}_i}$ is the gradient of E with respect to ${\theta}_i$, which is close to 0 because the pre-trained model is supposed to converge and the objective function was originally at its local minimum.
In the second term, $h_{ii} = \frac{\partial^2 E}{\partial {\theta}_i^2}$ is the element at row $i$ and column $i$ of second derivative Hessian matrix $\textbf{H}$.
The approximation assumes that $\Delta E$ caused by removing several parameters is the sum of $\Delta E$ caused by removing each parameter individually, so the third term $\frac{1}{2} \sum_{i\neq j}h_{ij}\theta_i \theta_j$ is neglected.
The fourth term $O(\theta_i^3)$ is apparently close to zero and should get neglected.
Therefore, the difference is approximated to only the second term $\frac{1}{2}\sum_{i} h_{ii} \theta_i^2$, where $\frac{1}{2} h_{ii} \theta_i^2$ effectively represent how much removing ${\theta}_i$ can affect the model outputs.

To improve fairness of the given DL model, our goal is to select which weights to drop so that we get close to the objectives of
\begin{equation}
    \min \Delta E_{c=0}, \quad \max \Delta E_{c=1},
\end{equation}
where $\Delta E_{c=0}$ and $\Delta E_{c=1}$ are the changes of objective function as mentioned in Equation~\ref{equ:taylor_exp} for two demographic groups, denoted as $c = 0$ and $c = 1$.
Therefore, by controlling the input sample $x$, we can now use two subsets divided by the sensitive attribute as input separately to evaluate how each weight of the matrix influences the output differently if removed for sparsity.
Once weight-wise importance results of a weight matrix on two groups are obtained, we can calculate fairness-aware ``scores'' for a weight $\theta_i$ in the matrix with a weighed subtraction:
\begin{equation}\label{equ:combined_score}
s_i = \frac{1}{2}h_{ii}^0 \theta_i^2 - \beta \cdot \frac{1}{2}h_{ii}^1 \theta_i^2 = \frac{1}{2}\theta_i^2 (h_{ii}^0 - \beta h_{ii}^1),
\end{equation}
where $\beta$ is a hyper-parameter controlling the trade-off between minimizing $\Delta E_{c=0}$ and maximizing $\Delta E_{c=1}$.
Here, for demonstration, we use $c = 0$ to represent the unprivileged group, and $c = 1$ for the privileged group.
We call $s_i$ as the fairness-aware score of ${\theta}_i$, where a smaller value represents a larger benefit when we remove it since a smaller $s_i$ means less contribution to $\Delta E_{c=0}$ and more to $\Delta E_{c=1}$, matching our goal mentioned above.
$h_{ii}^c$ is the Hessian element of ${\theta}_i$, measured on the scoring set of demographic group $c \in \{0, 1\}$.

\subsection{Workflow of FairLRF}

Figure~\ref{workflow} shows the entire workflow of FairLRF.
Starting from a pre-trained DL model, the conventional truncated SVD is applied first.
Then, we use two subsets of the training data, named scoring set, to conduct inference with the model, and collect the average Hessian results as the weight-wise importance for predicting on two demographic groups.
Next, use the weighed subtraction as Equation~\ref{equ:combined_score} goes to calculate weight-wise scores for fairness.
Finally, we conduct sparse SVD, yet with the guidance of fairness-aware scores.
In the example in Figure~\ref{workflow}, $sr \times D_{in}$ rows (marked with golden boxes) of $\hat{U}_{D_{in} \times k}$ with the minimum sum of weight-wise fairness scores are selected to drop the last $\lceil (1 - rr) \times k \rceil$ weights by zeroing them out.
Similarly, $\hat{V}_{k \times D_{out}}$ can be processed like this by column.
Then the entire FairLRF workflow is finished, leading to another advantage of FairLRF: it does not require any fine-tuning or retraining.
For FairLRF, the influence on the performance is mathematically minimized as long as hyper-parameters are set properly, because the nature of SVD determines that FairLRF always first drops weights of unitary matrices that correspond to the smallest singular values, i.e., with minimum influence on outputs.
By contrast, FairQuantize~\cite{guo2024fairquantize} is the state-of-the-art fairness-oriented quantization method, yet it needs fine-tuning after each step of quantization, or the accuracy would drop significantly.

\section{Experiments and Results}

\subsection{Experiment Setup}

\subsubsection{Datasets, Backbones, and Pre-processing}

The experiments are conducted on two datasets for image classification.
The first is CelebA~\cite{liu2015faceattributes}, which we intend to use as a general purpose image classification measurement.
It contains 202,599 face images of celebrities, each of which has 40 binary attributes (yes or no, or with or without), such as male (gender), hair color, attractive, big nose, etc.
There have been a variety of researches selecting different attributes as the target attribute and sensitive attribute.
In this paper, we follow \cite{xu2020investigating} to take ``Smiling'' as the target attribute (for classification) and ``Male'' as the sensitive attribute, since data distribution in terms of gender is imbalanced, leading to the potential performance bias (and so it is observed later).
The other dataset, Fitzpatrick-17k~\cite{groh2021evaluating,groh2022towards}, is a medical image datasets, representing our exploration of FairLRF usage in more realistic applications of DL models.
It is a dermatological disease dataset, containing 16,577 clinical images of 114 dermatological conditions for classification.
Images are categorized into 6 types of skin tones (marked as 1 to 6), from the lightest to the darkest.
We group them into light (1 to 3) and dark (4 to 6) groups, and use this binary skin tone attribute as the sensitive attribute.
Both datasets should add up to provide a comprehensive view of how FairLRf works compared with existing methods.

For pre-processing, each input image is first resized to $128 \times 128$, and cropped to $112 \times 112$.
If for training, data augmentation techniques like flipping and color distortion are randomly applied then.

As for the backbone models, VGG-11~\cite{simonyan2014very} is adopted as the backbone models.
The VGG architecture is known to have multiple fully connected layers to classify features extracted by convolutional layers.
Another reason is that it has also been frequently used in recent fairness-related works, including the fairness-aware model compression baseline methods used below.
Therefore, a pre-trained model is obtained on each dataset, denoted as vanilla in the baseline methods.

\subsubsection{Baselines}

On the one hand, as a tuned version of SVD, it is natural to compare with conventional SVD methods.
Therefore, truncated SVD and sparse SVD following the two strategies adopted in \cite{swaminathan2020sparse}, i.e., absolute weight and absolute activation, are tested below for comparison.
These methods are denoted in tables below as SLR-w and SLR-a, respectively.
It is noted though that mathematically, a conventional SVD-processed model is not involved in any approximation, so its behavior should be identical to the original one without any modification.
This is why we do not apply conventional SVD, an even more fundamental version of SVD, as a baseline method.

On the other hand, it is also crucial to compare with existing fairness-aware model compression solutions.
So we adopted FairPrune~\cite{wu2022fairprune} and FairQuantize~\cite{guo2024fairquantize} as baselines as well.
FairPrune utilizes pruning strategies, while FairQuantize is based on quantization principles.
These represent recent research efforts for fairness improvement with different approaches to model compression.

\begin{table}[ht] \centering \tiny
\caption{
Test set performance of methods with VGG-11 backbone model on CelebA.
The female and male gender groups are privileged and unprivileged, respectively.
All LRF configurations use $k=64$ and were fixed on validation data.
For SLR-w and SLR-a, ``fairness-opt. (non-deg.)'' minimizes $\max(\mathrm{EOpp},\mathrm{EOdd})$ over $0<\mathrm{subrank}<k$ with average precision no lower than vanilla on the validation set;
``FairLRF-aligned'' uses FairLRF's $(sr,rr)$.
Precision is group-wise macro precision;
``Avg.'' and ``Diff.'' are the unweighted group mean and absolute gap;
``C.R.'' is the theoretical model-size compression ratio.
All numbers are rounded to three decimal places, while average and difference calculations are conducted before rounding.
These conventions also apply to Table~\ref{table_fitzpatrick17k_vgg11}.
\label{table_celeba_vgg11}}
\resizebox{\linewidth}{!}{
\begin{tabular}{@{}lccccc@{}}
\toprule
\multirow{2}*{Method} & \multirow{2}*{Gender}
& \multirow{2}*{Precision $\uparrow$} & \multicolumn{2}{c}{Fairness Metrics} & \multirow{2}*{C.R. $\uparrow$} \\
\cline{4-5}
~ & ~ & ~ & EOpp $\downarrow$ & EOdd $\downarrow$ & ~ \\
\midrule
\multirow{4}*{Vanilla}
& Female               & 0.923 & \multirow{4}*{0.034} & \multirow{4}*{0.032} & \multirow{4}*{1.000x} \\
~ & Male               & 0.896 & ~ & ~ & ~ \\
~ & Avg.               & 0.909 & ~ & ~ & ~ \\
~ & Diff. $\downarrow$ & 0.026 & ~ & ~ & ~ \\
\midrule
\multirow{4}*{\shortstack[l]{Truncated SVD \\ ($k = 64$)}}
& Female               & 0.923 & \multirow{4}*{0.034} & \multirow{4}*{0.031} & \multirow{4}*{1.567x} \\
~ & Male               & 0.896 & ~ & ~ & ~ \\
~ & Avg.               & 0.909 & ~ & ~ & ~ \\
~ & Diff. $\downarrow$ & 0.027 & ~ & ~ & ~ \\
\midrule
\multirow{4}*{\shortstack[l]{SLR-w~\cite{swaminathan2020sparse} \\ \textit{fairness-opt. (non-deg.)} \\ ($k = 64, sr = 0.2, rr = 0.3$)}}
& Female               & 0.923 & \multirow{4}*{0.034} & \multirow{4}*{0.031} & \multirow{4}*{1.571x} \\
~ & Male               & 0.896 & ~ & ~ & ~ \\
~ & Avg.               & 0.909 & ~ & ~ & ~ \\
~ & Diff. $\downarrow$ & 0.026 & ~ & ~ & ~ \\
\midrule
\multirow{4}*{\shortstack[l]{SLR-w~\cite{swaminathan2020sparse} \\ \textit{FairLRF-aligned} \\ ($k = 64, sr = 0.2, rr = 0.25$)}}
& Female               & 0.923 & \multirow{4}*{0.034} & \multirow{4}*{0.031} & \multirow{4}*{1.572x} \\
~ & Male               & 0.896 & ~ & ~ & ~ \\
~ & Avg.               & 0.909 & ~ & ~ & ~ \\
~ & Diff. $\downarrow$ & 0.027 & ~ & ~ & ~ \\
\midrule
\multirow{4}*{\shortstack[l]{SLR-a~\cite{swaminathan2020sparse} \\ \textit{fairness-opt. (non-deg.)} \\ ($k = 64, sr = 0.75, rr = 0.25$)}}
& Female               & 0.923 & \multirow{4}*{0.034} & \multirow{4}*{0.031} & \multirow{4}*{1.584x} \\
~ & Male               & 0.896 & ~ & ~ & ~ \\
~ & Avg.               & 0.909 & ~ & ~ & ~ \\
~ & Diff. $\downarrow$ & 0.027 & ~ & ~ & ~ \\
\midrule
\multirow{4}*{\shortstack[l]{SLR-a~\cite{swaminathan2020sparse} \\ \textit{FairLRF-aligned} \\ ($k = 64, sr = 0.2, rr = 0.25$)}}
& Female               & 0.923 & \multirow{4}*{0.034} & \multirow{4}*{0.031} & \multirow{4}*{1.572x} \\
~ & Male               & 0.896 & ~ & ~ & ~ \\
~ & Avg.               & 0.909 & ~ & ~ & ~ \\
~ & Diff. $\downarrow$ & 0.027 & ~ & ~ & ~ \\
\midrule
\multirow{4}*{\shortstack[l]{FairPrune~\cite{wu2022fairprune} \\ (pruning ratio = 55\%) \\ ($\beta = \frac{3}{7}$)}}
& Female               & 0.922 & \multirow{4}*{0.034} & \multirow{4}*{0.032} & \multirow{4}*{1.000x} \\
~ & Male               & 0.896 & ~ & ~ & ~ \\
~ & Avg.               & 0.909 & ~ & ~ & ~ \\
~ & Diff. $\downarrow$ & 0.026 & ~ & ~ & ~ \\
\midrule
\multirow{4}*{\shortstack[l]{FairQuantize~\cite{guo2024fairquantize} \\ (quantization ratio = 15\%) \\ ($\beta = \frac{5}{9}$)}}
& Female               & 0.919 & \multirow{4}*{0.032} & \multirow{4}*{0.023} & \multirow{4}*{1.122x} \\
~ & Male               & 0.891 & ~ & ~ & ~ \\
~ & Avg.               & 0.905 & ~ & ~ & ~ \\
~ & Diff. $\downarrow$ & 0.028 & ~ & ~ & ~ \\
\midrule
\multirow{4}*{\shortstack[l]{FairLRF \\ ($k = 64, sr = 0.2, rr = 0.25$) \\ ($\beta=\frac{5}{9}$)}}
& Female               & 0.924 & \multirow{4}*{0.031} & \multirow{4}*{0.027} & \multirow{4}*{1.572x} \\
~ & Male               & 0.894 & ~ & ~ & ~ \\
~ & Avg.               & 0.909 & ~ & ~ & ~ \\
~ & Diff. $\downarrow$ & 0.029 & ~ & ~ & ~ \\
\bottomrule
\end{tabular}
}
\end{table}

\begin{table}[ht] \centering \tiny
\caption{Test-set performance of VGG-11 on Fitzpatrick-17k.
The dark and light skin tone groups are privileged and unprivileged, respectively.
\label{table_fitzpatrick17k_vgg11}}
\resizebox{\linewidth}{!}{
\begin{tabular}{@{}lccccc@{}}
\toprule
\multirow{2}*{Method} & \multirow{2}*{\shortstack[c]{Skin \\ Tone}}
& \multirow{2}*{Precision $\uparrow$} & \multicolumn{2}{c}{Fairness Metrics} & \multirow{2}*{C.R. $\uparrow$} \\
\cline{4-5}
~ & ~ & ~ & EOpp $\downarrow$ & EOdd $\downarrow$ & ~ \\
\midrule
\multirow{4}*{Vanilla}
& Light                & 0.503 & \multirow{4}*{0.281} & \multirow{4}*{0.141} & \multirow{4}*{1.000x} \\
~ & Dark               & 0.571 & ~ & ~ & ~ \\
~ & Avg.               & 0.537 & ~ & ~ & ~ \\
~ & Diff. $\downarrow$ & 0.068 & ~ & ~ & ~ \\
\midrule
\multirow{4}*{Truncated SVD}
& Light                & 0.503 & \multirow{4}*{0.264} & \multirow{4}*{0.132} & \multirow{4}*{1.558x} \\
~ & Dark               & 0.584 & ~ & ~ & ~ \\
~ & Avg.               & 0.544 & ~ & ~ & ~ \\
~ & Diff. $\downarrow$ & 0.081 & ~ & ~ & ~ \\
\midrule
\multirow{4}*{\shortstack[l]{SLR-w~\cite{swaminathan2020sparse} \\ \textit{fairness-opt. (non-deg.)} \\ ($k = 64, sr = 0.4, rr = 0.05$)}}
& Light                & 0.517 & \multirow{4}*{0.256} & \multirow{4}*{0.128} & \multirow{4}*{1.569x} \\
~ & Dark               & 0.586 & ~ & ~ & ~ \\
~ & Avg.               & 0.552 & ~ & ~ & ~ \\
~ & Diff. $\downarrow$ & 0.069 & ~ & ~ & ~ \\
\midrule
\multirow{4}*{\shortstack[l]{SLR-w~\cite{swaminathan2020sparse} \\ \textit{FairLRF-aligned} \\ ($k = 64, sr = 0.4, rr = 0.05$)}}
& Light                & 0.517 & \multirow{4}*{0.256} & \multirow{4}*{0.128} & \multirow{4}*{1.569x} \\
~ & Dark               & 0.586 & ~ & ~ & ~ \\
~ & Avg.               & 0.552 & ~ & ~ & ~ \\
~ & Diff. $\downarrow$ & 0.069 & ~ & ~ & ~ \\
\midrule
\multirow{4}*{\shortstack[l]{SLR-a~\cite{swaminathan2020sparse} \\ \textit{fairness-opt. (non-deg.)} \\ ($k = 64, sr = 0.1, rr = 0.9$)}}
& Light                & 0.502 & \multirow{4}*{0.266} & \multirow{4}*{0.133} & \multirow{4}*{1.559x} \\
~ & Dark               & 0.584 & ~ & ~ & ~ \\
~ & Avg.               & 0.543 & ~ & ~ & ~ \\
~ & Diff. $\downarrow$ & 0.081 & ~ & ~ & ~ \\
\midrule
\multirow{4}*{\shortstack[l]{SLR-a~\cite{swaminathan2020sparse} \\ \textit{FairLRF-aligned} \\ ($k = 64, sr = 0.4, rr = 0.05$)}}
& Light                & 0.546 & \multirow{4}*{0.269} & \multirow{4}*{0.135} & \multirow{4}*{1.569x} \\
~ & Dark               & 0.572 & ~ & ~ & ~ \\
~ & Avg.               & 0.559 & ~ & ~ & ~ \\
~ & Diff. $\downarrow$ & 0.027 & ~ & ~ & ~ \\
\midrule
\multirow{4}*{\shortstack[l]{FairPrune~\cite{wu2022fairprune} \\ (pruning ratio = 65\%) \\ ($\beta = \frac{1}{3}$)}}
& Light                & 0.506 & \multirow{4}*{0.256} & \multirow{4}*{0.129} & \multirow{4}*{1.000x} \\
~ & Dark               & 0.565 & ~ & ~ & ~ \\
~ & Avg.               & 0.535 & ~ & ~ & ~ \\
~ & Diff. $\downarrow$ & 0.059 & ~ & ~ & ~ \\
\midrule
\multirow{4}*{\shortstack[l]{FairQuantize~\cite{guo2024fairquantize} \\ (quantization ratio = 20\%) \\ ($\beta = 1$)}}
& Light                & 0.521 & \multirow{4}*{0.268} & \multirow{4}*{0.134} & \multirow{4}*{1.176x} \\
~ & Dark               & 0.586 & ~ & ~ & ~ \\
~ & Avg.               & 0.554 & ~ & ~ & ~ \\
~ & Diff. $\downarrow$ & 0.065 & ~ & ~ & ~ \\
\midrule
\multirow{4}*{\shortstack[l]{FairLRF \\ ($k = 64, sr = 0.4, rr = 0.05$) \\ ($\beta=\frac{1}{9}$)}}
& Light                & 0.599 & \multirow{4}*{0.210} & \multirow{4}*{0.105} & \multirow{4}*{1.569x} \\
~ & Dark               & 0.539 & ~ & ~ & ~ \\
~ & Avg.               & 0.569 & ~ & ~ & ~ \\
~ & Diff. $\downarrow$ & 0.060 & ~ & ~ & ~ \\
\bottomrule
\end{tabular}
}
\end{table}

\subsubsection{Metrics}

For predictive utility in the main comparison tables, in this paper, macro precision is reported on both demographic groups of each dataset, together with their unweighted average and absolute difference, as the representitive of accuracy metrics.
We especially pay attention to unweighted group-averaged macro precision (denoted as $P_{\mathrm{avg}}$) results below.
Overall accuracy is not chosen because it can be misleading on highly imbalanced datasets such as Fitzpatrick-17k.

For fairness metrics, we are going to report EOpp and EOdd as introduced above.
These metrics are different by definition, and they reflect the idea of fairness from different perspectives.
For instance, equalized opportunity is valuable for circumstances where fairness is expected basically within certain outcomes of the target attribute only, such as admission to a college or job position~\cite{hardt2016equality}.
Besides, there is no guarantee to be able to satisfy them together.
Therefore, which fairness metrics matter more strongly depend on users' preference as well as nature of tasks.

As for model compression, compression rates (denoted as C.R. in the tables) of sizes to store weights will be reported.
For one method, its compression rate is the total size to store weights of vanilla model divided by the total size to store weights of the model processed by this method.
It is noted that FairPrune is an unstructured pruning method, and weights are pruned discretely and irregularly from weight matrices, making it difficult to store, i.e., unable to actually compress model sizes, so its compression rate will always be 1.
FairQuantize compresses models by converting select weights from 32-bit floating point numbers to 8-bit fixed point integers, reducing the space to store quantized weights.
For LRF methods, weight matrices are indeed decomposed to smaller ones, and get even more sparse for SLR-w, SLR-a, and FairLRF.
Weights are removed from the model, so it gets compressed by LRF methods.
However, actual model file sizes are significantly influenced by too many overheads and compression-irrelevant factors, e.g., file format, whether to store weights only or with model architecture, potential extra information, etc.
Therefore, the compression rate results presented are calculated in theory only, and a higher compression rate does not necessarily leads to higher inference speed without proper handling.
However, it is still sufficient to provide insights into whether and how involved model compression methods work for their original purposes.
It is also worth noticing that the primary goal of the proposed FairLRF is to increase fairness of processed models instead of compression, so compression rates are less crucial than other two groups of metrics in this paper.

\subsection{Results on CelebA Dataset}

For this group of experiments, we apply LRF on the weight matrix of the last $4096 \times 4096$ fully connected layer, i.e., the last hidden layer, whose biases are kept unchanged.
Based on the preliminary experiments on validation sets (more details are available in the ablation study), $k = 64$ is used for truncated SVD, SLR-w, SLR-a, and our method reported below.

Table~\ref{table_celeba_vgg11} shows the results of our method and baseline methods working on VGG-11 models pre-trained on CelebA dataset.
While all methods manage to keep the average precision at approximately the same level as vanilla, our method reaches the best EOpp and the second best EOdd, slightly higher than FairQuantize.
If noticing the drop of avergae precision of FairQuantize, FairLRF actually is the best method in terms of fairness performance that does not reduce average precision in Table~\ref{table_celeba_vgg11}.
Compared to vanilla, our method maintains the same average precision while reducing EOpp by 10.6\% and EOdd by 14.9\% (calculated before rounding).
As for compression rates, due to the pretty low $k$, further sparsity of unitary matrices is not so significant as how truncated SVD reduces a great amount of weights, so SLR-w, SLR-a, and FairLRF are getting similar compression rates.

\subsection{Results on Fitzpatrick-17k Dataset}

As the previous group of experiments goes, LRF methods work on the last $4096 \times 4096$ fully connected layer.
Here, $k = 64$ is once again selected for LRF methods including truncated SVD, SLR-w, SLR-a, and our method.

Table~\ref{table_fitzpatrick17k_vgg11} shows the results of our method and baseline methods working on VGG-11 models pre-trained on Fitzpatrick-17k dataset.
Obviously, it is a much more complicated and challenging dataset compared with CelebA, as the accuracy performances of all methods are significantly lower than the ones in Table~\ref{table_celeba_vgg11}.
As the results demonstrate, FairLRF achieves the highest average precision, which is good and actully beyond our expectations.
What is more important, is that FairLRF also achieves both the best EOpp and EOdd.
Our method not only outperforms all other baseline methods on this tougher dataset, but also reaches significant 25.3\% reduction of EOpp and 25.5\% reduction of EOdd.
As for compression rates, similar to the CelebA case, the sparse rank-64 variants (SLR-w, SLR-a, and FairLRF) achieve pretty similar C.R.s (1.559x and 1.569x).

As we review results of two groups of experiments, FairLRF becomes even more promising given that only one fully connected layer is compressed by FairLRF to achieve the performance above, while FairPrune and FairQuantize work on the weights of the entire model.
What makes FairLRF even more valuable is that it requires no fine-tuning or re-training.
It is much more efficient for our method to produce processed models compared with FairPrune and FairQuantize that relies on re-training to regain lost accuracy performances.
This mostly thanks to the nature of SVD computation, which effectively retains more performances with less remaining weights, while leaving more space for potentially further optimization by sparse SVD and our work.

\begin{figure*}[t]
\centering
\includegraphics[width=0.7\textwidth]{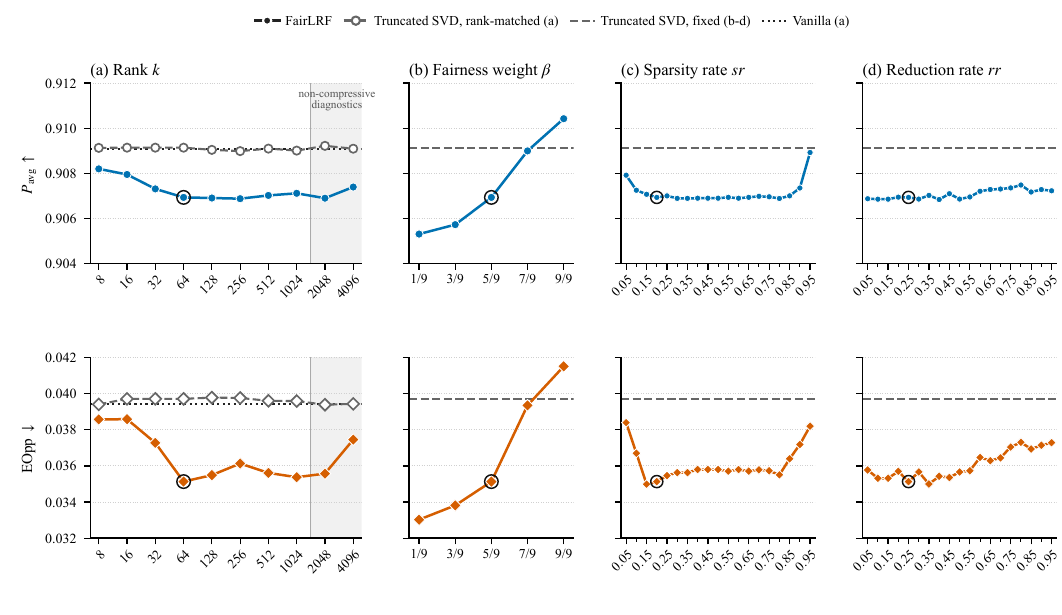}
\caption{
Hyper-parameter ablation study on CelebA dataset (validation set) and VGG-11 backbone model.
The top and bottom rows report average precision and EOpp, respectively.
Filled markers denote FairLRF.
In (a), the dotted horizontal line denotes a separately evaluated vanilla checkpoint, and the black circle marks the main-experiment rank $k = 64$.
(b) varies $\beta$ and displays the results with a common denominator of nine.
(c) and (d) are slices of Figure~\ref{fig:ablation_full_grid}: (c) varies $sr$ at $rr=0.25$, and (d) varies $rr$ at $sr=0.20$.
The latter three groups show the fixed rank-64 truncated SVD control as a gray horizontal line without markers.
Black circles mark the reported case in Table~\ref{table_celeba_vgg11}: $k = 64$, $\beta = \frac{5}{9}$, $sr = 0.20$, and $rr = 0.25$.
Connecting lines are visual guides only.
}
\Description{
Eight line plots show average precision and EOpp difference while varying rank, $\beta$, sparsity rate, and reduction rate.
}
\label{fig:ablation_hyperparameters}
\end{figure*}

\begin{figure*}[t]
\centering
\includegraphics[width=0.7\textwidth]{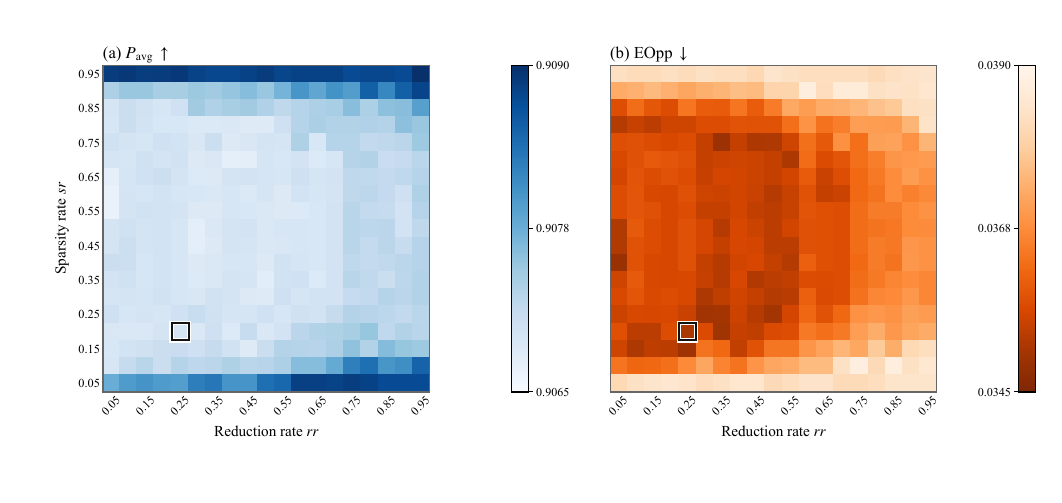}
\caption{
FairLRF performances over a uniformly spaced $19\times19$ FairLRF grid with $sr, rr\in\{0.05, 0.10, \ldots, 0.95\}$.
(a) reports average precision, and (b) reports EOpp.
Darker shading indicates higher $P_{\mathrm{avg}}$ in (a) and lower EOpp in (b);
the black outline marks the main-aligned cell $(sr, rr)=(0.20, 0.25)$.
The heatmap color limits show within-grid variation.
}
\Description{
Two dense 19-by-19 heatmaps show average precision and EOpp difference over sparsity and reduction rates from 0.05 through 0.95 at increments of 0.05.
}
\label{fig:ablation_full_grid}
\end{figure*}

\subsection{Ablation Study}

There are many possible hyper-parameter combinations for FairLRF, and the results reported above are only with the selected combinations.
The following experiments are actually conducted before the ones presented above, which sheds light on how hyper-parameters are properly determined to get the optimal results.
In addition, we also explore the possibility to apply FairLRF to more than just one layer.
To be clear, all experiments below are conducted before the main experiments in Table~\ref{table_celeba_vgg11} and Table~\ref{table_fitzpatrick17k_vgg11} to better search for appropriate hyper-parameter selections, so the following results are all reported on the validation set to avoid using the test set results to make decisions.
For simplicity, we only report average precision for accuracy, and EOpp for fairness.

\subsubsection{Influence of Ranks}

To begin with, Figure~\ref{fig:ablation_hyperparameters}(a) shows a group of experiments with only different ranks and everything else the same ($sr = 0.2$, $rr = 0.25$, $\beta = \frac{5}{9}$, target layer of the last hidden layer, etc.) that aligns with the FairLRF set in Table~\ref{table_celeba_vgg11}.
It compares every FairLRF point with a deterministic truncated SVD control of the same rank.
We sweep ten powers of two from $k=8$ to $k = 4096$.

The first thing that can be easily noticed is the slightly lower average precision of FairLRF in exchange for lower EOpp compared with truncated SVD across all different rank values.
Apparently this effect comes from our fairness-aware sparse process of SVD matrices.
Among these ranks with different trade-off situations, there is not any visible trend that can tell which ranks are always better.
The only thing we do know, though, is that significantly small ranks lead to less difference between FairLRF and truncated SVD in terms of both average precision and EOpp.
It is as expected: very small ranks produce small decomposed matrices, meaning that there are few elements that can be potentially dropped by FairLRF, making it less capable of influencing the model performance.
The black circle mark reveals how we choose the reported result of FairLRF in Table~\ref{table_celeba_vgg11}: now that precision decrease is not very significant, we choose the rank that potentially improves fairness the most, which is 64 in our case.

\subsubsection{Influence of $\beta$}

Figure~\ref{fig:ablation_hyperparameters}(b) shows FairLRF performance change with different $\beta$ values, with truncated SVD marked with a gray horizontal line for comparison, which does not get tuned by $\beta$.
It can be found that $\beta$ shows a very clear trade-off between accuracy and fairness: for this specific scenario, higher $\beta$ in the given range result in both better average precision and worse EOpp.
In other words, there is not any one-for-all method to guide the choice of the most proper $\beta$ for an arbitrary backbone model and dataset combination.
Instead, it mostly depends on specific conditions that FairLRF works with.
The fact that optimal choices of $\beta$ for FairLRF in Table~\ref{table_celeba_vgg11} and Table~\ref{table_fitzpatrick17k_vgg11} are different is another evidence as well.
Therefore, our choice of $\frac{5}{9}$ is more like some sort of balance between two dimensions, and it is hard to tell whether there may be even better $\beta$ options given the potential impacts from other hyper-parameters.

\subsubsection{Sparsity rate $sr$ and reduction rate $rr$}

Figure~\ref{fig:ablation_hyperparameters}(c) and Figure~\ref{fig:ablation_hyperparameters}(d) shows how $sr$ and $rr$ tunes FairLRF in comparison with truncated SVD, which can be considered as an extreme case of $sr = 0$ or $rr = 1$ for FairLRF.
For (c), we control $rr = 0.25$ the same as the reported result in Table~\ref{table_celeba_vgg11} and only look for the impact of $sr$ alone.
Similarlly, for (d), $sr = 0.2$, and only $rr$ changes.
It does show the same overall trend mentioned above: FairLRF leads to both worse average precision and better EOpp, in that all four curves go below the reference line of truncated SVD.
However, similar to $rank$, there is no straightforward rules of choosing $sr$ and $rr$, e.g., it is basically impossible to tell from these results that either smaller or bigger ones work always better.
As a result, it is of vital importance to conduct sufficient preliminary experiments to search for optimal hyper-parameter sets so that FairLRF can produce the strongest power.
Figure uses a heatmap to further demonstrate this:

\begin{figure}[t]
\centering
\includegraphics[width=0.87\linewidth]{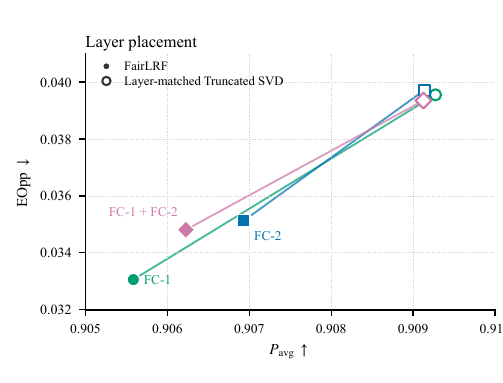}
\caption{
FairLRF and truncated SVD results of average precision and EOpp on FC-1, FC-2, or both layers of VGG-11 backbone model on CelebA.
Each line joins FairLRF to its layer-matched truncated SVD control.
}
\Description{
A paired scatter plot compares FairLRF and layer-matched truncated SVD controls for the first fully connected layer, the second fully connected layer, and their joint application.
}
\label{fig:ablation_layers_actual}
\end{figure}

\subsubsection{Layer Selection Also Matters}

VGG-11 has three fully connected layers after convolutional layers in total, denoted as FC-1, FC-2, and FC-3 in this paper.
In a conventional VGG-11 model, after convolutional layers (and the pooling layer), FC-1, FC-2, and FC-3 connects consecutively, followed by softmax.
In other words, all LRF experiments above work on FC-2.
Naturally, it is interesting to see whether the optimization also applies to FC-1 given the exactly the same setups, and how combining them together can change results.
It is noted that FC-3 is excluded here, as its weight matrix size is $4096 \times C_{out}$.
For CelebA, $C_{out} = 2$.
This makes the highest possible rank comes down to merely 2, leaving very limited room for all SVD-based methods to work with, including FairLRF.

As Figure~\ref{fig:ablation_layers_actual} goes, the layer-matched truncated SVD controls yield $(P_{\mathrm{avg}},\mathrm{EOpp})=(0.9093,0.0395)$ for FC-1, $(0.9091,0.0397)$ for FC-2, and $(0.9091,0.0393)$ jointly.
FairLRF moves these operating points to $(0.9056,0.0331)$, $(0.9069,0.0351)$, and $(0.9062,0.0348)$, respectively.
Since we expect higher average precision and lower EOpp, lower right points on this figure are better.
However, all these points clearly form a trade-off band lying from higher right to lower left on the figure, once again show the same trend for FairLRF to gain fairness while losing accuracy simultaneously compared to truncated SVD.
While applying the same hyper-parameters means that FC-1 and FC-1 + FC-2 points here do not guarantee to reach the optimal results, this figure is sufficient to demonstrate that choosing different layer(s) also influences how FairLRF performs.

\section{Conclusion}

In this work, we propose FairLRF, a framework that utilizes SVD, a model compression method, to improve fairness.
It evaluates the importance of weights for reducing unfairness based on the well-validated observation that weights in conventional DL models are often redundent and contribute to model prediction differently.
Extensive experiments show that FairLRF outperforms conventional SVD methods as well as existing fairness-oriented model compression methods, achieving an optimal balance between accuracy and fairness on two datasets with different complexities.
In addition, an ablation study systematically measures the influence of hyper-parameter selections, providing basic guidance for potential usages of FairLRF in the future.

{
    \small
    \bibliographystyle{ieeenat_fullname}
    \bibliography{main}

@inproceedings{guo2024fairquantize,
  title={Fairquantize: Achieving fairness through weight quantization for dermatological disease diagnosis},
  author={Guo, Yuanbo and Jia, Zhenge and Hu, Jingtong and Shi, Yiyu},
  booktitle={International Conference on Medical Image Computing and Computer-Assisted Intervention},
  pages={329--338},
  year={2024},
  organization={Springer}
}

@article{ferrara2024fairness,
  title={Fairness-aware machine learning engineering: how far are we?},
  author={Ferrara, Carmine and Sellitto, Giulia and Ferrucci, Filomena and Palomba, Fabio and De Lucia, Andrea},
  journal={Empirical software engineering},
  volume={29},
  number={1},
  pages={9},
  year={2024},
  publisher={Springer}
}

@incollection{dastin2022amazon,
  title={Amazon scraps secret AI recruiting tool that showed bias against women},
  author={Dastin, Jeffrey},
  booktitle={Ethics of data and analytics},
  pages={296--299},
  year={2022},
  publisher={Auerbach Publications}
}

@article{obermeyer2019dissecting,
  title={Dissecting racial bias in an algorithm used to manage the health of populations},
  author={Obermeyer, Ziad and Powers, Brian and Vogeli, Christine and Mullainathan, Sendhil},
  journal={Science},
  volume={366},
  number={6464},
  pages={447--453},
  year={2019},
  publisher={American Association for the Advancement of Science}
}

@article{du2020fairness,
  title={Fairness in deep learning: A computational perspective},
  author={Du, Mengnan and Yang, Fan and Zou, Na and Hu, Xia},
  journal={IEEE Intelligent Systems},
  volume={36},
  number={4},
  pages={25--34},
  year={2020},
  publisher={IEEE}
}

@article{dunkelau2019fairness,
  title={Fairness-aware machine learning},
  author={Dunkelau, Jannik and Leuschel, Michael},
  journal={An extensive overview},
  pages={1--60},
  year={2019}
}

@inproceedings{wu2022fairprune,
  title={Fairprune: Achieving fairness through pruning for dermatological disease diagnosis},
  author={Wu, Yawen and Zeng, Dewen and Xu, Xiaowei and Shi, Yiyu and Hu, Jingtong},
  booktitle={Medical Image Computing and Computer Assisted Intervention--MICCAI 2022: 25th International Conference, Singapore, September 18--22, 2022, Proceedings, Part I},
  pages={743--753},
  year={2022},
  organization={Springer}
}

@inproceedings{kong2024achieving,
  title={Achieving fairness through channel pruning for dermatological disease diagnosis},
  author={Kong, Qingpeng and Chiu, Ching-Hao and Zeng, Dewen and Chen, Yu-Jen and Ho, Tsung-Yi and Hu, Jingtong and Shi, Yiyu},
  booktitle={International Conference on Medical Image Computing and Computer-Assisted Intervention},
  pages={24--34},
  year={2024},
  organization={Springer}
}

@article{swaminathan2020sparse,
  title={Sparse low rank factorization for deep neural network compression},
  author={Swaminathan, Sridhar and Garg, Deepak and Kannan, Rajkumar and Andres, Frederic},
  journal={Neurocomputing},
  volume={398},
  pages={185--196},
  year={2020},
  publisher={Elsevier}
}

@article{hardt2016equality,
  title={Equality of opportunity in supervised learning},
  author={Hardt, Moritz and Price, Eric and Srebro, Nati},
  journal={Advances in neural information processing systems},
  volume={29},
  year={2016}
}

@article{lebedev2018speeding,
  title={Speeding-up convolutional neural networks: A survey},
  author={Lebedev, Vadim and Lempitsky, Victor},
  journal={Bulletin of the Polish Academy of Sciences. Technical Sciences},
  volume={66},
  number={6},
  pages={799--811},
  year={2018},
  publisher={Polska Akademia Nauk. Czasopisma i Monografie PAN}
}

@article{cheng2017survey,
  title={A survey of model compression and acceleration for deep neural networks},
  author={Cheng, Yu and Wang, Duo and Zhou, Pan and Zhang, Tao},
  journal={arXiv preprint arXiv:1710.09282},
  year={2017}
}

@article{lecun1989optimal,
  title={Optimal brain damage},
  author={LeCun, Yann and Denker, John and Solla, Sara},
  journal={Advances in neural information processing systems},
  volume={2},
  year={1989}
}

@inproceedings{liu2015faceattributes,
  title = {Deep Learning Face Attributes in the Wild},
  author = {Liu, Ziwei and Luo, Ping and Wang, Xiaogang and Tang, Xiaoou},
  booktitle = {Proceedings of International Conference on Computer Vision (ICCV)},
  month = {December},
  year = {2015}
}

@inproceedings{xu2020investigating,
  title={Investigating bias and fairness in facial expression recognition},
  author={Xu, Tian and White, Jennifer and Kalkan, Sinan and Gunes, Hatice},
  booktitle={European Conference on Computer Vision},
  pages={506--523},
  year={2020},
  organization={Springer}
}

@article{simonyan2014very,
  title={Very deep convolutional networks for large-scale image recognition},
  author={Simonyan, Karen and Zisserman, Andrew},
  journal={arXiv preprint arXiv:1409.1556},
  year={2014}
}

@inproceedings{groh2021evaluating,
  title={Evaluating deep neural networks trained on clinical images in dermatology with the fitzpatrick 17k dataset},
  author={Groh, Matthew and Harris, Caleb and Soenksen, Luis and Lau, Felix and Han, Rachel and Kim, Aerin and Koochek, Arash and Badri, Omar},
  booktitle={Proceedings of the IEEE/CVF Conference on Computer Vision and Pattern Recognition},
  pages={1820--1828},
  year={2021}
}

@article{groh2022towards,
  title={Towards transparency in dermatology image datasets with skin tone annotations by experts, crowds, and an algorithm},
  author={Groh, Matthew and Harris, Caleb and Daneshjou, Roxana and Badri, Omar and Koochek, Arash},
  journal={arXiv preprint arXiv:2207.02942},
  year={2022}
}

@inproceedings{chiu2023toward,
  title={Toward fairness through fair multi-exit framework for dermatological disease diagnosis},
  author={Chiu, Ching-Hao and Chung, Hao-Wei and Chen, Yu-Jen and Shi, Yiyu and Ho, Tsung-Yi},
  booktitle={International Conference on Medical Image Computing and Computer-Assisted Intervention},
  pages={97--107},
  year={2023},
  organization={Springer}
}

@article{guo2024hardware,
  title={Hardware design and the fairness of a neural network},
  author={Guo, Yuanbo and Yan, Zheyu and Yu, Xiaoting and Kong, Qingpeng and Xie, Joy and Luo, Kevin and Zeng, Dewen and Wu, Yawen and Jia, Zhenge and Shi, Yiyu},
  journal={Nature Electronics},
  volume={7},
  number={8},
  pages={714--723},
  year={2024},
  publisher={Nature Publishing Group UK London}
}

@INPROCEEDINGS{7995254,
  author={Vasudevan, Aravind and Anderson, Andrew and Gregg, David},
  booktitle={2017 IEEE 28th International Conference on Application-specific Systems, Architectures and Processors (ASAP)},
  title={Parallel Multi Channel convolution using General Matrix Multiplication},
  year={2017},
  volume={},
  number={},
  pages={19-24},
  doi={10.1109/ASAP.2017.7995254}}

@ARTICLE{9650846,
  author={Trusov, Anton V. and Limonova, Elena E. and Nikolaev, Dmitry P. and Arlazarov, Vladimir V.},
  journal={IEEE Access},
  title={p-im2col: Simple Yet Efficient Convolution Algorithm With Flexibly Controlled Memory Overhead},
  year={2021},
  volume={9},
  number={},
  pages={168162-168184},
  doi={10.1109/ACCESS.2021.3135690}}
}

\end{document}